\documentclass[twoside,11pt]{article}

\usepackage[preprint]{jmlr2e}
\usepackage{booktabs}
\usepackage{pifont}
\usepackage{listings}

\ShortHeadings{ModSSC: A Modular Framework for Semi-Supervised Classification}{Melvin Barbaux}
\firstpageno{1}

\title{ModSSC: A Modular Framework for Semi-Supervised Classification on Heterogeneous Data}
\author{%
  \name Melvin Barbaux \email melvin.barbaux@math.u-bordeaux.fr\\
  \name Samia Boukir \email samia.boukir@math.u-bordeaux.fr \\
  \addr Univ. Bordeaux, CNRS, Bordeaux INP, IMB, UMR 5251, F-33400 Talence, France
}

\begin{document}

\maketitle

\begin{abstract}

\indent Semi-supervised classification leverages both labeled and unlabeled data to improve predictive performance, but existing software support remains fragmented across methods, learning settings, and data modalities. We introduce ModSSC, an open source Python framework for inductive and transductive semi-supervised classification designed to support reproducible and controlled experimentation.

ModSSC provides a modular and extensible software architecture centered on reusable semi-supervised learning components, stable abstractions, and fully declarative experiment specification. Experiments are defined through configuration files, enabling systematic comparison across heterogeneous datasets and model backbones without modifying algorithmic code.

ModSSC~1.0.0 is released under the MIT license with full documentation and automated tests, and is available at \url{https://github.com/ModSSC/ModSSC}. The framework is validated through controlled experiments reproducing established semi-supervised learning baselines across multiple data modalities.
\end{abstract}


\section{Introduction}
Semi-supervised classification is essential when labels are costly. It combines a small labeled set with a large pool of unlabeled samples to improve generalization beyond purely supervised approaches.

Despite extensive methodological progress, software support for semi-supervised classification remains fragmented. Implementations are dispersed across independent repositories, often tightly coupled to specific algorithms, data modalities, or experimental settings. Switching between methods or backbones typically requires reimplementing data loading, training loops, and evaluation protocols. This fragmentation undermines reproducibility and fair comparison, since empirical differences may stem from inconsistent pipelines rather than intrinsic algorithmic properties \citep{Oliver2018RealisticSSL}.

Recent advances further exacerbate these issues. Large pretrained models can be adapted to semi-supervised classification via fine tuning on limited labeled data combined with the exploitation of unlabeled samples \citep{Chen2020BigSSL}. Evaluating such approaches in a controlled and comparable manner requires software infrastructures that decouple semi-supervised strategies from model architectures, datasets, and training mechanics.

To address these limitations, we propose ModSSC, a software framework that does not introduce new semi-supervised algorithms, but instead focuses on software abstractions, modularity, and reproducible experimentation across semi-supervised learning settings. ModSSC is not a fixed benchmark suite, but an extensible infrastructure intended to support reusable experimentation and strengthen comparative analysis rather than leaderboard-style evaluation.

\paragraph{Contributions.}
\begin{itemize}
    \item A fully declarative experiment specification separating datasets, sampling schemes, model backbones, training procedures, and evaluation protocols from purely algorithmic code.
    \item Stable abstract interfaces that decouple semi-supervised learning strategies from classifiers and machine learning libraries, enabling reuse across architectures, modalities, and execution environments.
    \item A unified software abstraction covering both inductive and transductive semi-supervised learning, including graph-based methods, with a consistent evaluation pipeline.
    \item A maintained and tested implementation layer for a broad set of well-established semi-supervised classification methods, integrated under a common and stable API.
\end{itemize}

\section{Framework Overview}

ModSSC is organized around a software design that separates experiment specification, semi-supervised learning strategies, and execution logic. Data preparation, model backbones, and semi-supervised methods are treated as independent components, reducing coupling between algorithmic choices and implementation details.

The software is structured in two complementary layers. A core library implements reusable building blocks for semi-supervised classification, including data abstractions, learning strategies, and model interfaces. An orchestration layer composes these components into executable experimental workflows for systematic evaluation and comparison.

The core library can be used independently of the orchestration layer, which is intended to support controlled experimental studies rather than to define a fixed benchmark. This separation is enforced through stable abstract interfaces defining explicit contracts between data representations, model backbones, and semi-supervised strategies.

The orchestration layer associates each experiment with a deterministic identifier derived from its configuration and manages the caching of intermediate artifacts such as datasets, sampling splits, graphs, and learned representations. This mechanism ensures that repeated runs of identical configurations yield comparable results at the level of experimental protocol and evaluation metrics.

\section{Implementation and Usage}

ModSSC~1.0.0 is an open source Python package publicly available at \url{https://github.com/ModSSC/ModSSC}. The framework is distributed through the Python Package Index (\texttt{pip install modssc}). All experiments reported in this paper correspond to the tagged release \texttt{v1.0.0}, with explicitly version-bounded dependencies specified in the project configuration files.

ModSSC relies on a small set of widely used Python scientific and machine learning libraries, including NumPy for numerical computation, scikit-learn for classical learning components, PyTorch as primary deep learning backend, and PyTorch Geometric for graph-based transductive models, with optional dependencies activated only when the corresponding backends are used.

The codebase is covered by automated unit tests and continuously validated through integration tests, static type checking, and style enforcement. The package is accompanied by documentation including API references, tutorial notebooks, and executable examples covering both inductive and transductive semi-supervised workflows.

Experiments are specified through a single YAML configuration file defining the task, dataset, sampling scheme, training procedure, and semi-supervised method, and are executed from the command line on CPU or GPU depending on the selected backbone, using at most one GPU per model. This execution model supports systematic replication of experimental protocols without modifying source code.

The current release exposes 50 semi-supervised classification methods, corresponding to distinct algorithmic formulations as defined in their original publications, and excluding architectural variants or minor implementation ablations. The exact catalogue corresponding to version~1.0.0 is documented in Appendix~\ref{app:methods}.

When official reference implementations are available, the corresponding ModSSC implementations are derived directly from those codebases and adapted to the framework architecture without altering the underlying algorithms. For methods without official implementations, the code follows the pseudo-code and algorithmic descriptions provided in the original publications, with adaptations limited to interface integration and software modularization.

ModSSC prioritizes experimental reliability and reduction of engineering overhead rather than per-run algorithmic speedups. Wall-clock performance depends primarily on the chosen backbone models and execution environment.

\section{Scope and Design Choices}

ModSSC is intentionally scoped to semi-supervised classification. Other learning tasks such as regression, clustering, or representation learning are not addressed directly, except when used as intermediate components within a classification pipeline.

The framework does not aim to provide a fixed or exhaustive benchmark. Instead, it prioritizes extensibility and controlled experimentation, allowing users to reproduce experimental protocols and explore new combinations of datasets, models, and semi-supervised strategies.

ModSSC guarantees reproducibility at the level of software artifacts, experimental configurations, and evaluation protocols. It does not aim to guarantee exact numerical reproduction of results reported in original methodological papers, due to differences in hardware, numerical backends, and non-deterministic training components inherent to many modern learning pipelines.

\section{Related Work}

Several open source libraries provide partial software support for semi-supervised learning, but typically focus on specific algorithmic families or learning settings. We consider only methods exposed through a documented and stable public API, excluding experimental scripts, ablation variants, and architecture-specific reimplementations. A quantitative comparison is reported in Table~\ref{tab:ssl-comparison}.

LAMDA SSL \citep{Jia2023LAMDA} integrates statistical and deep semi-supervised algorithms within scikit-learn style pipelines, but exposes semi-supervised strategies as task-specific learners rather than reusable components. The Unified Semi-Supervised Benchmark (USB) \citep{Wang2022USB} targets benchmarking of modern deep semi-supervised methods under standardized protocols, but focuses on inductive deep classifiers within a single backend. GraphLearning \citep{graphlearning} concentrates on graph-based transductive learning and does not naturally extend to non-graph modalities or inductive pipelines.

Beyond quantitative coverage, existing libraries differ fundamentally in their software abstractions. Most frameworks expose semi-supervised methods as task-specific learners tightly coupled to a given backend or data modality, whereas ModSSC considers semi-supervised strategies as reusable components that can be combined with heterogeneous backbones and datasets under a unified execution model. This design enables controlled cross-setting experimentation that is difficult to achieve within benchmark-oriented or modality-specific toolkits.

\begin{table}[ht!]
\centering
\caption{Comparison of semi-supervised learning toolkits. Numbers refer to semi-supervised classification algorithms exposed through a documented and stable public API, excluding experimental scripts and minor variants.}
\label{tab:ssl-comparison}
\small
\setlength{\tabcolsep}{4pt}
\renewcommand{\arraystretch}{1.15}
\begin{tabular}{@{}lcccc@{}}
\toprule
 & \textbf{LAMDA} & \textbf{USB} & \textbf{GraphLearning} & \textbf{ModSSC} \\
\midrule
Inductive algorithms    & 26 & 14 & 0  & 28 \\
Transductive algorithms & 2  & 0  & 14 & 22 \\
\midrule
Data modalities         & 4  & 3  & 2  & 5  \\
\bottomrule
\end{tabular}
\end{table}

\section{Conclusion}

We presented ModSSC, an open-source Python framework designed to support reproducible research in semi-supervised classification across inductive and transductive learning settings. By decoupling semi-supervised strategies from model backbones and execution logic, the framework enables controlled comparisons across datasets and computational environments. Beyond replication, ModSSC provides a software foundation for studying the transferability of semi-supervised learning strategies across data modalities, supporting systematic reuse beyond their original experimental settings.

ModSSC is extensible and open to community development. Its modular architecture lowers the engineering barrier for researchers and practitioners seeking to deploy semi-supervised methods in new domains.

\bibliography{ModSSC_Semi-Supervised_Classification_Framework}

\appendix
\section{Catalogue of Implemented Methods}
\label{app:methods}

This appendix documents the set of semi-supervised classification methods included in ModSSC version~1.0.0.

\begin{table}[ht!]
\centering
\small
\setlength{\tabcolsep}{8pt}
\renewcommand{\arraystretch}{1.2}
\caption{Classical methods integrated into ModSSC.}
\label{tab:ssl_methods_classical}
\begin{tabular}{@{}p{0.48\linewidth}p{0.48\linewidth}@{}}
\toprule
\textbf{Inductive} & \textbf{Transductive} \\
\midrule
ADSH \citep{Guo2022AdaptiveThresholding} & Dynamic Label Propagation \citep{Wang2016DynamicLabelPropagation} \\
Co-Training \citep{Blum1998CoTraining} & Graph Mincuts \citep{Blum2001GraphMincuts} \\
Democratic Co-Learning \citep{Zhou2004DemocraticCoLearning} & GraphHop \citep{Xie2023GraphHop} \\
S4VM \citep{Li2011S4VM} & Label Propagation \citep{Zhu2003LabelPropagation} \\
Self Training \citep{Yarowsky1995} & Label Spreading \citep{Zhou2003LabelSpreading} \\
SETRED \citep{Li2005SETRED} & Laplace Learning \citep{Zhu2003GFHF} \\
Tri-Training \citep{Zhou2005TriTraining} & Lazy Random Walk \citep{Zhou2004RandomWalks} \\
TSVM \citep{Joachims1999TSVM} & p-Laplace Learning \citep{Flores2022LpSSL} \\
 & Poisson Learning \citep{Calder2020PoissonLearning} \\
 & Poisson MBO \citep{Calder2020PoissonLearning} \\
\bottomrule
\end{tabular}
\end{table}

\begin{table}[ht!]
\centering
\small
\setlength{\tabcolsep}{8pt}
\renewcommand{\arraystretch}{1.2}
\caption{Neural methods integrated into ModSSC.}
\label{tab:ssl_methods_neural}
\begin{tabular}{@{}p{0.48\linewidth}p{0.48\linewidth}@{}}
\toprule
\textbf{Inductive} & \textbf{Transductive} \\
\midrule
AdaMatch \citep{Berthelot2022AdaMatch} & APPNP \citep{Klicpera2019APPNP} \\
CoMatch \citep{Li2021CoMatch} & ChebNet \citep{Defferrard2016ChebNet} \\
DASO \citep{Oh2022DASO} & GAT \citep{Velickovic2018GAT} \\
DeFixMatch \citep{Schmutz2023DeFixMatch} & GCN \citep{Kipf2017GCN} \\
Deep Co-Training \citep{Qiao2018DeepCoTraining} & GCNII \citep{Chen2020GCNII} \\
FixMatch \citep{Sohn2020FixMatch} & GraFN \citep{Lee2022GraFN} \\
FlexMatch \citep{Zhang2021FlexMatch} & GRAND \citep{Feng2020GRAND} \\
FreeMatch \citep{Wang2023FreeMatch} & GraphSAGE \citep{Hamilton2017GraphSAGE} \\
Mean Teacher \citep{Tarvainen2017MeanTeacher} & H-GCN \citep{Hu2019HGCN} \\
Meta Pseudo Labels \citep{Pham2021MetaPseudoLabels} & N-GCN \citep{AbuElHaija2020NGCN} \\
MixMatch \citep{Berthelot2019MixMatch} & Planetoid \citep{Yang2016Planetoid} \\
Noisy Student \citep{Xie2020NoisyStudent} & SGC \citep{Wu2019SGC} \\
Pi-Model \citep{Laine2016PiModel} & \\
Pseudo Label \citep{Lee2013PseudoLabel} & \\
SimCLRv2 \citep{Chen2020BigSSL} & \\
SoftMatch \citep{Chen2023SoftMatch} & \\
Temporal Ensembling \citep{Laine2016PiModel} & \\
Tri-Net \citep{Chen2018TriNet} & \\
UDA \citep{Xie2020UDA} & \\
VAT \citep{Miyato2019VAT} & \\
\bottomrule
\end{tabular}
\end{table}

\end{document}